\pdfoutput=1
\documentclass[11pt]{article}

  \usepackage[final]{other/acl}
  \usepackage{times}
  \usepackage{booktabs}
  \usepackage{latexsym}
  \usepackage[T1]{fontenc}
  \usepackage[utf8]{inputenc}
  \usepackage{microtype}
  \usepackage{tikz-dependency}
  \usepackage{hyperref}
  \usepackage{xcolor}
  \usepackage{tabulary}

\newcommand\blfootnote[1]{%
  \begingroup
  \renewcommand\thefootnote{}\footnote{#1}%
  \addtocounter{footnote}{-1}%
  \endgroup
}
 
\title{\textsc{Mukayese}: Turkish NLP Strikes Back}

\def \subsubsectionspacing {5pt}

\author{Ali Safaya$^{\dagger,}$\footnotemark{}  \ \ Emirhan Kurtuluş$^{\ddag}$ \ \ Arda Göktoğan$^{\S}$ \ \ Deniz Yuret$^{\dagger}$ \\ \\
  $^{\dagger}$ KUIS AI Center, Koç University \ $^{\dagger}$ Computer Engineering Department, Koç University \\
  $^{\ddag}$ Cağaloğlu Anadolu Lisesi, Istanbul \ $^{\S}$ Computer Engineering Department, Bilkent University}

\begin{document}
  \maketitle
  \blfootnote{Corresponding author: \texttt{asafaya19@ku.edu.tr}}
  \begin{abstract}

Having sufficient resources for language X lifts it from the \textit{under-resourced} languages class, but not necessarily from the \textit{under-researched} class. In this paper, we address the problem of the absence of organized benchmarks in the Turkish language. We demonstrate that languages such as Turkish are left behind the state-of-the-art in NLP applications. As a solution, we present \textsc{Mukayese}, a set of NLP benchmarks for the Turkish language that contains several NLP tasks. We work on one or more datasets for each benchmark and present two or more baselines. Moreover, we present four new benchmarking datasets in Turkish for language modeling, sentence segmentation, and spell checking. All datasets and baselines are available under: \url{https://github.com/alisafaya/mukayese}

\end{abstract}

  \section{Introduction}
\label{sec:introduction}

Although some human languages, such as Turkish, are not classified as under-resourced languages, only a few research communities are working on them \cite{joshi-etal-2020-state}. As a result, they are left behind in developing state-of-the-art systems due to the lack of organized benchmarks and baselines. In this study, we aim to address this gap for the Turkish language with \textsc{Mukayese} (Turkish word for "comparison/benchmarking"), an extensive set of datasets and benchmarks for several Turkish NLP tasks.

We survey several tasks in Turkish NLP and observe an absence of organized benchmarks and research. We demonstrate how the lack of benchmarks affects under-studied languages such as Turkish and how it can keep the state of research behind the state-of-the-art of NLP. We accomplish this by presenting state-of-the-art baselines that outperform previous work significantly. We believe that \textsc{Mukayese} will set a basis for boosting NLP research for Turkish. Therefore, we encourage research communities from other under-studied languages to follow a similar path.

In our work on \textsc{Mukayese}, we study seven NLP tasks in the Turkish language. We evaluate available datasets in Turkish for these tasks and describe the process of creating four new datasets for tasks that do not have accessible datasets. Furthermore, in addition to evaluating existing methods, we provide at least two baseline models/methods per task. More details are enlisted in Table \ref{tab:intro-tasks}.

Our overall contribution to Turkish NLP can be summarized as the following: (a) Set of seven organized benchmarks for NLP. (b) Four new datasets in Turkish for language modeling, sentence segmentation, as well as spellchecking and correction. (c) Dataset splits for fair benchmarking. (d) Several replicable baselines for each task. (e) Benchmarking state-of-the-art methods on Turkish.

Moreover, Mukayese is a part of the Turkish Data Depository (TDD) project\footnote{\url{https://tdd.ai}}. The main goal of TDD is collecting and organizing Turkish Natural Language Processing NLP resources and providing a research basis for Turkish NLP.

The rest of the paper is organized as follows: We review similar efforts in Section \ref{sec:related}. Then, we advert to benchmarks and NLP in Section \ref{sec:benchmarks}. Next, we give a background on the Turkish language resources in Section \ref{sec:background}. We explain the approach we follow for each task in Section \ref{sec:approach}, and we provide dataset details, evaluation results, and explain the baselines for each task in Section \ref{sec:tasks}.

\begin{table*}[t]
\centering
\small
\begin{tabular}{llll}
\toprule[1.5pt]
\normalsize \textsc{Task} &  \normalsize \textsc{Datasets} & \normalsize  \textsc{Metrics} &  \normalsize \textsc{Baselines}  \vrule depth 2ex height 2.5ex width 0pt \\ \midrule
{\sc Language Modeling } &
{\sc  \begin{tabular}[c]{@{}l@{}}- \textbf{trnews-64}\\- \textbf{trwiki-67}\end{tabular}  } &
{\sc  \begin{tabular}[c]{@{}l@{}}- Bits-per-char\\- Perplexity\end{tabular} } &
{\sc  \begin{tabular}[c]{@{}l@{}}- Adapt. Trans. \\- SHA-RNN\end{tabular} } \\  \midrule
{\sc Machine Translation } &
{\sc  \begin{tabular}[c]{@{}l@{}}- Wmt-16\\- MuST-C\end{tabular} } &
{\sc  - BLEU  } &
{\sc  \begin{tabular}[c]{@{}l@{}}- ConvS2S\\- Transformer\\- mBART50\end{tabular} } \\ \midrule
{\sc Named-Entity Recognition } &
{\sc  \begin{tabular}[c]{@{}l@{}}- WikiAnn\\- Milliyet-Ner\end{tabular} } &
{\sc  - CoNLL F1 } &
{\sc  \begin{tabular}[c]{@{}l@{}}- Bilstm-Crf\\- Bert \\- Bert-Crf\end{tabular} }\\ \midrule
{\sc Sentence Segmentation } &
{\sc  - \textbf{trseg-41} } &
{\sc  - Segment F1-Score } &
{\sc  \begin{tabular}[c]{@{}l@{}}- spaCy\\- Punkt\\- Ersatz\end{tabular} } \\ \midrule
{\sc Spellchecking \& Correction } &
{\sc  - \textbf{trspell-10} } &
{\sc  \begin{tabular}[c]{@{}l@{}}- F1-Score\\- Accuracy\end{tabular} } &
{\sc  \begin{tabular}[c]{@{}l@{}}- Zemberek\\- Hunspell\end{tabular} } \\ \midrule
{\sc Summarization } &
{\sc  - Mlsum } &
{\sc  \begin{tabular}[c]{@{}l@{}}- Rouge-L\\- Meteor\end{tabular} } &
{\sc  \begin{tabular}[c]{@{}l@{}}- Transformer\\- mBart50\\- mT5\end{tabular} } \\ \midrule 
{\sc Text Classification } &
{\sc  \begin{tabular}[c]{@{}l@{}}- Offenseval \\- \textit{News-Cat} \end{tabular} }  &
{\sc  \begin{tabular}[c]{@{}l@{}}- F1-Score \end{tabular} } &
{\sc  \begin{tabular}[c]{@{}l@{}}- Bilstm\\- CNN Text\\- Bert\end{tabular} } \\ \bottomrule[1.5pt]
\end{tabular}
\caption{List of the NLP Tasks we work on for the Turkish language in \textsc{Mukayese}. We list the datasets, metrics, and baselines we use for each task. New datasets presented in this paper are marked in \textbf{bold}, and ones for which we present train/test splits are marked in \textit{italic}.}
\label{tab:intro-tasks}
\vspace{-5pt}
\end{table*}
  \section{Related Work}
\label{sec:related}

In this section, we discuss efforts similar to ours. We give an overview of efforts on building multilingual benchmarks, and we mention some of the monolingual benchmarks as well.

\vspace{20pt}


There exist various endeavors at building multilingual benchmarks. One example for this is \textsc{Xtreme} \cite{hu-etal-2020-xtreme}, a multilingual benchmark containing 40 different languages and nine different tasks. These tasks include Classification, Named Entity Recognition (NER), and Question Answering (QA). However, most of these datasets are created by translating existing English datasets manually or automatically. Therefore, they have limitations and cannot be utilized to build a research basis in a specific language.


There are several benchmarks for NLP tasks for both low-resource and high-resource languages when it comes to monolingual benchmarks.
\citet{duh-etal-2020-benchmarking} proposes a benchmark for two low-resourced African languages on Neural Machine Translation (NMT), namely Somali and Swahili. Similarly, there are efforts to build benchmarks for high-resource but under-studied languages such as \textsc{Alue} benchmark for Arabic \cite{seelawi-etal-2021-alue}, and \textsc{Klej} benchmark for Polish \cite{rybak-etal-2020-klej}. Both benchmarks focus on Natural Language Understanding (NLU). Most of these benchmarks have public leaderboards to disseminate studies in NLP for their languages.

While most of previous benchmarks focus on one task such as NLU or NMT, \textsc{Mukayese} covers a comprehensive set of NLP tasks with seven different benchmarks on a variety of tasks. The reasoning behind this is to catalyze the research of Turkish NLP, and encourage research in all NLP applications.

\section{Benchmarks and NLP}
\label{sec:benchmarks}

Following the research on NLP over the years, we can observe how datasets and benchmarks are fundamental. In this section, we discuss the importance of benchmarks for the progress of NLP. 

Benchmarks are very essential for measuring the progress of NLP. For instance, the SQuAD dataset \cite{rajpurkar-etal-2016-squad} is used to examine the progress of English Question Answering, and GLUE \cite{wang-etal-2018-glue}, SuperGLUE \cite{wang-etal-2019-superglue} provide benchmarks for English Language Understanding.

Such progress has been enabled by the existence of benchmarks, which allowed for fair and meaningful comparison, and showed if there is room for improvement. In addition, organized benchmarks and datasets enable the research community to make progress with minimal amount of domain knowledge.

This is especially important when it comes to languages with fewer speakers, and research communities are more likely to contribute when such organized tasks are presented \cite{plumed-etal-2021}. Thus, this is essential if we want to include other communities in the development of under-resourced and under-studied languages.

However, there are several things to keep in mind when dealing with benchmarks and leaderboards. Such leaderboards should be created transparently, and the results need to be evaluated with all factors taken into account. Some of these factors are model size, energy efficiency, and generalization \cite{linzen-2020-accelerate}. Otherwise, we can run into the risk of these leaderboards resulting in inefficient and non-robust models. \citet{ethayarajh-jurafsky-2020-utility} describe a few limitations of current leaderboards and suggest practices to mitigate these limitations.

We take these practices into account and present the benchmarks of \textsc{Mukayese}. We provide more details about our methodology in Section \ref{sec:approach}.

  \section{Background on Turkish}
\label{sec:background}


The Turkish language has distinctive characteristics compared to well-studied languages in the literature, such as English, Spanish, and German. Due to its agglutinative morphological nature, Turkish nouns can produce more than 100 inflected forms, while verbs can produce even more \cite{oflazer-saraclar-2018-turkishnlp}. Therefore benchmarks designed for English are not necessarily applicable for Turkish.

\begin{figure}[h!]
    \centering
    \includegraphics[width=0.48\textwidth]{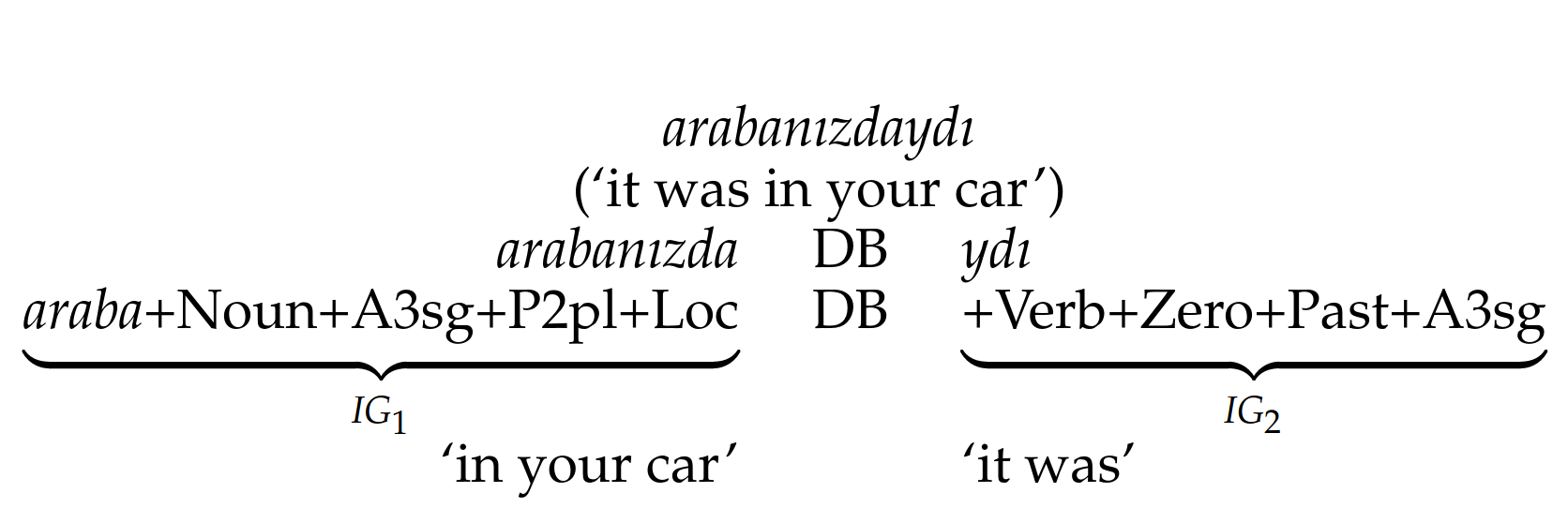}
    \caption{An example of a word constituting multiple inflectional groups \cite{eryiugit-etal-2008-dependency}.}
    \label{fig:IG}
\end{figure}
\vspace{-5pt}

Unlike many other languages, a single word can constitute multiple different inflectional groups. An example is displayed in Figure \ref{fig:IG}. We provide more details on the features of the Turkish language in Appendix \ref{app:turkish}.

There are several attempts at constructing comprehensive sets of resources and evaluation for Turkish. \citet{sak-et-al-2008} introduced a morphological parser, and a morphological disambiguator accompanied by a web corpus. More recently, \citet{eryigit-2014-itu} proposed an online Turkish NLP Pipeline, which includes Normalization, Tokenization, Morphological Analysis, NER, and Syntactic Parsing.

However, among previously proposed methods and datasets, none are presented in a comparative way. This study aims to make a comprehensive inventory of different tools, corpora, and evaluation measures for the Turkish language. Such inventory may be used for researchers and practitioners who are looking for tools and datasets for Turkish NLP.
  \section{Methodology}
\label{sec:approach}

In \textsc{Mukayese}, we focus on under-researched tasks of NLP in the Turkish language. After defining the task and assessing its importance, we construct the following three key elements for each benchmark:

\vspace{\subsubsectionspacing}
\textbf{Datasets } are the first element to consider when it comes to a benchmark. We define the minimum requirements of a benchmark dataset as follows: (i) accessible with reasonable size. (ii) Satisfactory quality. (iii) A publicly shareable, compliant applicable regulations (GDPR licensing).

We chose the dataset sizes in a task-specific manner, unless used in a few-shot setting, benchmarks with small datasets will lack generalizability, and models trained on them might suffer from overfitting. On the other hand, training models on enormous datasets might be costly and inefficient \cite{ethayarajh-jurafsky-2020-utility}.

Another feature to assess is the quality of the dataset. A manually annotated dataset with a low Interannotator Agreement (IAA) rate is not suitable for benchmarking. Moreover, to build a generalizable benchmark, we need to consider using a dataset representing the general domain. For instance, sentence segmentation methods of editorial texts do not work on user-generated content such as social media posts, as we show in Subsection \ref{subsec:sent-segmentation}.

\vspace{\subsubsectionspacing}
\textbf{Metrics } are the second element of benchmarks. We need to decide on one or more evaluation metrics to evaluate and compare methodologies. In order to do so, we have to answer the following questions:
(a) Does this metric measure what our task aims to do?
(b) How well does it correlate with human judgment?
(c) Are there any issues/bugs to consider in these metrics? (For example, using accuracy to measure performance on an unbalanced set does not give a representative idea of model performance).

\vspace{\subsubsectionspacing}
\textbf{Baselines } are the final element of benchmarking. In order to characterize the performance characteristics of different methodologies, it is better to diversify our baselines as much as possible. For instance, we can compare pretrained vs. non-pretrained approaches, rule-based systems vs. trained systems, or unsupervised vs. supervised models.

\section{Tasks}
\label{sec:tasks}

We provide benchmarks in the form of Datasets, Metrics, Baselines triplets for each of the following NLP tasks:

%
%

\subsection{Language Modeling}
\label{subsec:languagemodeling}

Auto-regressive language modeling is a generative process, which focuses on modeling the probability $P(X)$ of a text sequence of $n$ tokens, where $X = (x_1, x_2, ..., x_n)$, and $P(X) = \prod_{i=1}^{n}P(x_i|x_{<i})$. This type of language modeling is known as Auto-regressive (AR) or causal language modeling. The main objective of the model is to learn to estimate the probability of a given text sequence.

In our work, we focus on neural approaches for this task \cite{bengio-etal-2000-neurallm}, where we present two new benchmarking corpora for AR language modeling and report the results of two different baseline models.

\vspace{\subsubsectionspacing}
\textbf{Datasets \ \ }
We present two different corpora for AR language modeling, namely \textsc{trnews-64} and \textsc{trwiki-67}, along with their train/validation/test splits. These corpora are presented in a similar fashion to enwik8 \cite{hutter-2006-enwik8} and WikiText \cite{merity-etal-2017-wiki103} English corpora. We provide statistics about these corpora in Table \ref{tab:lm-datasets} in Appendix \ref{app:lm}.

\vspace{1pt}
\textbf{\textsc{trwiki-67} } is a language modeling corpus that contains 67 million words of raw Turkish Wikipedia articles. We extracted this corpus from a recent Turkish Wikipedia dump\footnote{https://dumps.wikimedia.org/trwiki/20210720/: accessed on 20 July 2021.} using WikiExtractor \cite{attardi-2015-wikiextractor}. Additionally, further pre-processing was applied to get rid of the redundant text. Only the articles' raw text and titles were kept and presented in their cased format (with no upper/lower case transformations). 

Due to the agglutinative nature of the Turkish language, most of the words are derived by combining one or more suffixes with one of the roots \cite{oflazer-saraclar-2018-turkishnlp}. To make use of this attribute of the Turkish language, we train a sentencepiece unigram model \cite{kudo-2018-subword} with a vocabulary size of 32K, only using the training split of the corpus. Although we advise using the tokenized version of this corpus to encourage reproducibility, we provide a raw version of this corpus that can be utilized as a benchmark for language modeling tasks on a character, subword, or word level.

\vspace{1pt}
\textbf{\textsc{trnews-64} } is another language modeling corpus that contain 64 million words of news columns and articles that was retrieved from TS Timeline Corpus \cite{sezer-2017-tscorpus}. It can be utilized as a benchmark for modeling long-range dependencies in the Turkish language, as it contains relatively long documents (See Table \ref{tab:lm-datasets}). This corpus consists of a mix of news articles collected from different journals about various domains and topics. Since \textsc{trnews-64} is intended for language modeling on character level, articles were lightly pre-processed, and no further tokenization was applied. We provide more details about \textsc{trnews-64} and \textsc{trwiki-67} in Appendix \ref{app:lm}.

\vspace{\subsubsectionspacing}
\textbf{Metrics \ \ } 
Language models are trained on minimizing the negative log-likelihood (\textsc{Nll}) of the training set, and their performance is measured based on how well they can generalize on the test set:
\begin{equation}
	\textsc{Nll}(X_{test}) = -\frac{1}{n} \sum_{i=1}^{n}log\ p_\theta(x_i|x_{test}{<i})
	\label{eq:nll}
\end{equation}
Word or sub-word level language models are evaluated using the word perplexity (\textsc{Ppl}) metric, a derivative of \textsc{Nll}. On the other hand, character language models are evaluated using entropy-based Bits-per-character (\textsc{Bpc}) metric, which is also another derivative of \textsc{Nll} \cite{chip-2019-evaluationlm}. We consider \textsc{Ppl} for the evaluation of models on \textsc{trwiki-67}, and \textsc{Bpc} for \textsc{trnews-64}. Note that lower is better for both metrics. 

We note that \textsc{Ppl} needs to be computed with the same count of tokens, otherwise it needs to be normalized (See Appendix \ref{app:lm}). Moreover, models considered to be evaluated using either of these corpora, are meant to have no training data other than that corpus' training split.

\begin{table}[h]
	\centering
	\small
	\begin{tabular}{lcccc}
		\toprule[1.5pt]
		& \multicolumn{2}{c}{\textsc{trwiki-67}} & \multicolumn{2}{c}{\textsc{trnews-64}} \\ \cmidrule(lr){2-5}
		& \textsc{\#Param}        & \textsc{Ppl}         & \textsc{\#Param}        & \textsc{Bpc}         \\ \midrule 
		\textsc{Adap.Trans} & 92M            & 14.64        & 38M            & 1.024        \\ 
		\textsc{Sha-Rnn}     & 87M            & \textbf{12.54}        & 53M            & \textbf{0.938}        \\ \bottomrule[1.5pt]
	\end{tabular}
	\caption{Results of language modeling baseline models, with their no of parameters. Perplexity (\textsc{Ppl}) is reported for \textsc{trwiki-67}, and Bits-per-char (\textsc{Bpc}) for \textsc{trnews-64}, on their test sets.}
	\label{tab:lm-results}
	\vspace{-10pt}
\end{table}

\vspace{\subsubsectionspacing}
\textbf{Baselines \ \ }
We consider two baseline models of different families. The first one is Single Headed Attention - RNN (\textsc{Sha-Rnn}) \cite{merity-2019-sharnn}, which is a Recurrent Neural Network-based language model, and the second is Adaptive Transformer (\textsc{Adap.Trans}) \cite{sukhbaatar-etal-2019-adaptive}, which is based on Transformer architecture \cite{vaswani-etal-2017-attention}. 
We choose these models for two main reasons. First, we want to compare models from different families (RNNs vs. Transformers). Second, compared to their counterparts such as \cite{lei-2021-attention, dai-etal-2019-transformer}, these models represent the state-of-the-art when it comes to the ratio of performance to the training cost and the number of parameters. For more details on the training refer to Appendix \ref{app:lm}.

In Table \ref{tab:lm-results}, we provide the results of these models, which we train and evaluate separately on \textsc{trwiki-67} and \textsc{trnews-64} corpora (See Table \ref{tab:lm-datasets} for more details on the splits of each corpus).

Note that even though we follow the same architectural settings for character-level and subword-level modeling, different tokenization algorithms of \textsc{trwiki-67} (subword-level) and \textsc{trnews-64} (character-level) lead to different vocabulary sizes, which leads to a difference in the number of parameters.

Unlike the case for the English language \cite{merity-2019-sharnn}, \textsc{Sha-Rnn} performed better than Adaptive Transformer for both of the presented Turkish corpora. This implies the necessity of establishing such benchmarks for other languages as well. We leave investigating this feature for future research.

\subsection{Machine Translation}
\label{subsec:machinetranslation}

Machine translation is the problem of translating a piece of text from one language to another. Over the years, neural machine translation models have become dominant, especially in low resource settings, benefiting from transfer learning \cite{zoph-etal-2016-transfer}. In this work, we focus on evaluating neural machine translation models for translation between English and Turkish languages. We provide the results of three different baselines on two datasets.

\vspace{\subsubsectionspacing}
\textbf{Datasets \ \ } The first dataset we evaluate is the Turkish-English subset of \textsc{Wmt-16}\footnote{http://www.statmt.org/wmt16/}, it consists of manually translated Turkish-English sentence pairs. The second one is the Turkish-English subset of Multilingual Speech Translation Corpus (\textsc{MuST-C}) \cite{digangi-etal-2019-mustc}. For details on the split refer to Table \ref{tab:mt-datasets} in Appendix \ref{app:mt}.

\vspace{\subsubsectionspacing}
\textbf{Metrics \ \ }  We evaluate our models on the relevant test sets for translation in both directions. We utilize \textsc{Bleu} Score \cite{papineni-etal-2002-bleu} for the assessment of translation quality.

\begin{table}[ht]
	\centering
    \small
	\begin{tabular}{lcccc}
		\toprule[1.5pt]
            & \multicolumn{2}{c}{\textsc{Wmt-16}} & \multicolumn{2}{c}{\textsc{MuST-C}} \\ 
& tr-en & en-tr & tr-en & en-tr \\ \midrule
            \multicolumn{2}{l}{\textit{from scratch}} & & & \\
            \citet{stahlberg-etal-2018-simple} & 19.17 & 13.61 & - & - \\
            \textsc{ConvS2S} (180M) & 13.22 & 12.78 & 21.79 & 13.3 \\ 
            \textsc{Trans.} (58M) & 17.29 & 15.72 & 27.01 & 15.52 \\ \midrule
            \multicolumn{2}{l}{\textit{pre-trained}} & & &  \\
            \textsc{mBart50} (680M) & \textbf{24.17} & \textbf{18.54} & \textbf{32.97} & \textbf{19.61} \\
         \bottomrule[1.5pt]
	\end{tabular}
	\caption{BLEU scores of machine translation baselines. Results are provided for translations in both directions.}
	\label{tab:mt-results}
	\vspace{-10pt}
\end{table}

\vspace{\subsubsectionspacing}
\textbf{Baselines \ \ } In this task, we train three different models. First, we train a \textsc{Transformer} \cite{vaswani-etal-2017-attention} with the same settings for the encoder and the decoder parts, where we use 6 layers, with 4 attention heads each, and hidden size of 512.
Second, we utilize the Convolutional sequence-to-sequence \textsc{ConvS2S} model \cite{gehring-etal-2017-convs2s} following the same settings. The last model is mBART 50 \cite{tang-etal-2020-mbart50}, a multilingual model pre-trained on 50 different languages, which we fine-tune for each dataset separately. 

In Table \ref{tab:mt-results}, we present \textsc{Bleu} score of the models on each translation dataset in both directions. The benefit of pre-training can be seen in the case of \textsc{mBART50}, where it outperforms the counterparts that we train from scratch.
Additionally, we compare our work to the results reported by \citet{stahlberg-etal-2018-simple} on \textsc{Wmt-16}. Their model is based on fusing language model decoding into seq2seq model with dot-attention \cite{luong-etal-2015-effective}.



\subsection{Named-Entity Recognition (NER)}
\label{subsec:ner}

We include the Named-Entity Recognition (NER) task in our set of benchmarks, as it has an essential role in NLP applications. In this task, words representing named-entities are detected in the text input and assigned one of the predefined named-entity classes such as \textit{Person} or \textit{Location} \cite{chinchor-robinson-1998-appendix}. We benchmark three different models on two NER datasets for Turkish and compare our work with previous work.

\vspace{\subsubsectionspacing}
\textbf{Datasets \ \ }
The first dataset we use is \textsc{Milliyet-Ner} \cite{tur-etal-2003-milliyet}, which is a set of manually, annotated news articles from the Turkish Milliyet news resource\footnote{https://www.milliyet.com.tr/}. The second is the Turkish subset of the semi-automatically annotated Cross-lingual NER dataset \textsc{WikiANN} or (PAN-X) \cite{pan-etal-2017-cross}, which consists of Turkish Wikipedia articles. Both datasets have three entity classes as shown in Table \ref{tab:ner-datasets} in Appendix \ref{app:ner}.

\vspace{\subsubsectionspacing}
\textbf{Metrics \ \ }
Following previous work on Turkish NER \cite{yeniterzi-2011-exploiting, seker-eryigit-2012-initial}, we report the CoNLL F-1 metric \cite{tjong-kim-sang-2002-introduction} to assess our NER baselines. CoNLL F-1 counts a named entity as correct, only if it is an exact match of the corresponding entity in the ground truth.

\begin{table}[h]
	\centering
	\small
	\begin{tabular}{lllll}
		\toprule[1.5pt]
		& \textsc{Milliyet} & \textsc{WikiAnn} \\ \midrule
		\cite{yeniterzi-2011-exploiting} & 91.56 & - \\ 
		\cite{seker-eryigit-2012-initial} & 91.94 & - \\ 
		\cite{gungor-etal-2018-rnnner} & 93.37 & - \\ \midrule
		\textsc{BiLSTM-CRF} & 95.54 & \textbf{93.8}        \\
		\textsc{BerTURK} & 95.31 & 92.82 \\
		\textsc{BerTURK-CRF} & \textbf{96.48} & 93.07 \\ \bottomrule[1.5pt]
	\end{tabular}
	\caption{Evaluation results (CoNLL $F_1$) of NER models on test sets.}
	\label{tab:ner-results}
	\vspace{-15pt}
\end{table}

\vspace{\subsubsectionspacing}
\textbf{Baselines \ \ } 
We train three different baseline models for this task. One with no pre-trained embeddings, which utilizes bi-directional Long Short Term Memory with Conditional Random Fields (\textsc{BiLSTM-CRF}) \cite{panchendrarajan-amaresan-2018-bidirectional}. The remaining two models employ pre-trained representations from BERT \cite{devlin-etal-2019-bert}. In one of the models, we investigate the benefit of adding a CRF layer on top of BERT. As for the pre-trained BERT model, we use \textsc{BerTURK} base, which is pre-trained on a large Turkish corpus \cite{schweter-2020-berturk}.

In Table \ref{tab:ner-results}, we provide the evaluation results (CoNLL $F_1$) for the three baselines on both datasets' test sets. Additionally, we compare our results with previous work of \cite{yeniterzi-2011-exploiting, seker-eryigit-2012-initial, gungor-etal-2018-rnnner} on the \textsc{Milliyet-Ner} dataset. We note that CoNLL $F_1$ of human performance on Turkish NER is expected to be in the range of 98-99\% \cite{tur-etal-2003-milliyet}.

\subsection{Sentence Segmentation}
\label{subsec:sent-segmentation}

Sentence segmentation is the task of detecting sentence boundaries in a given article. Despite its fundamental place in the NLP pipelines, sentence segmentation attracts little interest. Common approaches are rule-based systems that rely on cues such as punctuation marks and capital letters \cite{jurafsky-2018-speech}.

\vspace{\subsubsectionspacing}
\textbf{Datasets \ \ } We present \textsc{trseg-41}, a new sentence segmentation dataset for Turkish. This dataset consists of 300 sampled scientific abstracts from \cite{ozturk-etal-2014-turkishcorpus}, 300 curated news articles from \textsc{trnews-64}, and a set of 10K tweets. For the scientific abstracts, our sampling rationale is to maximize the number of abbreviations that reduce the accuracy of the rule-based approaches. As for the news subset, we maximize the length of documents and the number of proper nouns. In the Twitter subset, we balance the number of multi/single sentence tweets, and preprocess the tweets by replacing all URLs with \texttt{http://some.url}, and all user mentions with \texttt{@user}. 

A single annotator labels the sentence boundaries of the data samples. We present two dataset splits, one for training and development and one for testing and benchmarking. The statistics of the splits can be found in Table \ref{tab:segm-datasets} in Appendix \ref{app:sentenceseg}.

Applying sentence segmentation to user-generated content such as social media posts or comments can be quite challenging. To simulate such difficult cases and expose the weaknesses of rule-based methods, we create another version of \textbf{\textsc{trseg-41}} where we artificially corrupt the boundaries of sentences. This is done by randomly converting sentences to lowercase or uppercase with 50\% probability, or by removing all punctuation marks with 50\% probability.


\vspace{\subsubsectionspacing}
\textbf{Metrics \ \ }  Our evaluation procedure is based on the metrics F1 score, Precision, Recall for each segment. Unlike \cite{wicks-post-2021-unified}, we evaluate our models on the entire test set, without removing sentences with ambiguous boundaries. Furthermore, in order to highlight the gap in performance, we cross-evaluate our systems on the original and corrupted set.

\begin{table}[ht]
	\centering
    \small
	\begin{tabular}{lllll}
		\toprule[1.5pt]
		              & \textsc{F1-score} & \textsc{Precision}  & \textsc{Recall} \\ \midrule
		\textsc{spaCy}   & 0.74 / 0.37 & 0.76 / 0.48   & 0.72 / 0.30 \\ \midrule
		\multicolumn{2}{l}{\textit{Training (Original)}} & & \\
		\textsc{Ersatz}  & \textbf{0.89} / \textbf{0.40}  & \textbf{0.98} / 0.51   & 0.81 / \textbf{0.33} \\
		\textsc{Punkt}   & 0.87 / 0.39  &  0.88 / \textbf{0.52}    & \textbf{0.86} / 0.32  \\ \midrule
		\multicolumn{2}{l}{\textit{Training (Corrupted)}} & & \\
		\textsc{Ersatz}  & \textbf{0.88} / \textbf{0.40} & \textbf{0.97} / \textbf{0.51}   & 0.81 / \textbf{0.33} \\
		\textsc{Punkt}   & 0.85 / 0.39 & 0.86 / 0.50    & \textbf{0.84} / 0.31  \\
         \bottomrule[1.5pt]
	\end{tabular}
	\caption{Results of sentence segmentation baselines. Metrics are reported for both corrupted and clean versions of the test set in the \textsc{Original / Corrupted} format. }
	\label{tab:segm-results}
	\vspace{-15pt}
\end{table}

\vspace{\subsubsectionspacing}
\textbf{Baselines \ \ } For this task, we employ three methods as baseline models. \textsc{Ersatz}, a context-based approach that relies on supervised training \cite{wicks-post-2021-unified}, the unsupervised \textsc{Punkt} tokenizer \cite{kiss-strunk-2006-unsupervised}, and \textsc{spaCy} Sentencizer tool \cite{montani-etal-2021-spacy}. While \textsc{Ersatz} utilizes the Transformer \cite{vaswani-etal-2017-attention} architecture, spaCy Sentencizer is a rule-based sentence boundary detector, whereas Punkt Tokenizer relies on an unsupervised training approach.

We experiment with these models on four different training and testing set combinations, where we train using the original and corrupted training sets separately and test on both test sets. Results are presented in Table \ref{tab:segm-results}. In all settings, \textsc{spaCy Sentencizer} is outperformed by its trained counterparts. Among the baselines, \textsc{Ersatz} performed the best. Our experiments show that deep learning models are more robust to corruption in the data.

Please refer to Appendix \ref{app:sentenceseg} for dataset creation process and samples, and an analysis on the behaviour of our baselines.

\subsection{Spellchecking and Correction}
\label{subsec:spelling}

Spellcheckers are among the most widely used NLP tools. The basic task is to check for misspellings in an input and suggest a set of corrections. Different methods can be employed for error correction, such as looking up words that minimize the edit distance from a dictionary or utilizing probabilistic models with N-grams to suggest the most likely correct word based on the context \cite{jurafsky-2018-speech}. Due to the complexity of the Turkish Morphology, it is possible to derive over a hundred of words from one verb \cite{oflazer-saraclar-2018-turkishnlp}. This makes the spellchecking task quite challenging. Hence, we focus on contextless (single word) spellchecking and correction as a start, and leave in-context spellchecking for future work.

We present a new benchmarking dataset for contextless spellcheckers and a computationally efficient and accurate dictionary for Turkish.

\vspace{\subsubsectionspacing}
\textbf{Datasets \ \ } We present \textbf{\textsc{trspell-10}}, a dataset of 10K words, for benchmarking spellchecking and correction. The dataset consists of tuples of input and correct (gold) words.

To create this dataset, we randomly sample 8500 Turkish words from the TS Corpus Word List \cite{sezer-2013-tscorpus,sezer-2017-tscorpus}. We create artificial misspellings by applying random insertions, deletions, and substitutions on 65\% of the words, where we apply at most two operations on the same word. The remaining 35\% of the words are unchanged. Moreover, we add 1K random foreign words, and 500 randomly generated word-like character sequences.

As a quality check of these artificial misspellings, given a list of corrupted words, we ask our annotators to provide us a list of suggestions up to 10 suggestions per word. Their suggestion lists had the gold output 91\% of the time.

\vspace{\subsubsectionspacing}
\textbf{Metrics \ \ } We evaluate spellcheckers' ability to detect misspellings using the macro-averaged F1-Score metric. Additionally, we evaluate their spell correction accuracy (SCA) based on the suggestions provided for misspelled words.

\begin{table}[ht]
	\centering
	\small
	\begin{tabular}{lcc}
		\toprule[1.5pt]
    	                & \textsc{SCA}  & \textsc{F1} \\ \midrule
		\textsc{Hunspell-tr} \cite{hrzafer-2017-unspell} & 25.52 & 86.52  \\
		\textsc{Zemberek} \cite{akin-dundar-2007-zemberek} & 62.12 & 96.56 \\ \midrule
		\textsc{Our Hunspell}     &	\textbf{71.72} & \textbf{99.62} \\ \bottomrule[1.5pt]
	\end{tabular}
	\caption{Spell correction accuracy (SCA) and macro-averaged F1 scores of spellchecking methods on \textsc{trspell-10}.}
	\label{tab:spellchecking-results}
	\vspace{-15pt}
\end{table}

\vspace{\subsubsectionspacing}
\textbf{Baselines \ \ } We take advantage of the agglutinative nature of the Turkish language by developing a Hunspell-based \cite{tron-etal-2005-hunmorph} dictionary for Turkish. Using a list of 4M words we filter from Web crawls and Turkish corpora, we optimize the splits that minimize the size of the root dictionary and the affix list.

We compare this dictionary to \textsc{Hunspell-tr} \cite{hrzafer-2017-unspell} another Hunspell-based Turkish dictionary, and to \textsc{Zemberek} spellchecker \cite{akin-dundar-2007-zemberek}, which is designed based on morphological features of the Turkish language. As shown in Table \ref{tab:spellchecking-results}, our dictionary surpasses other baselines in terms of both error correction accuracy and error detection ability.

For dataset creation process and samples, please refer to Appendix \ref{app:spellcheck}.


\subsection{Summarization}
\label{subsec:summarization}

Abstractive text summarization is the task of generating a short description (summary) of an article (longer text). Formally, given a sequence of tokens (input article) $X = (x_1, x_2, ..., x_n)$ and its summary $Y = (y_1, y_2, ..., y_m)$, the main task is to model the conditional probability: $P(Y|X) = \prod_{i=1}^{m}P(y_i|y_{<i}, X)$. 

For this task, we work on the Multi-lingual Summarization (\textsc{Mlsum}) dataset \cite{scialom-etal-2020-mlsum} and present state-of-the-art summarization results for Turkish.

\vspace{\subsubsectionspacing}
\textbf{Datasets \ \ } \textsc{Mlsum} is a multi-lingual dataset for abstractive summarization. This dataset consists of a large set of crawled news articles with their abstracts in multiple languages. We focus on the Turkish subset of \textsc{Mlsum}.

We removed 4378 duplicated instances and 12 overlapping instances among the splits while assessing the dataset's quality. Further details in Appendix \ref{app:sum}.

\vspace{\subsubsectionspacing}
\textbf{Metrics \ \ } To assess the quality of the generated summaries, we use the N-gram co-occurrence-based \textsc{Rouge-l} \cite{lin-2004-rouge} and \textsc{Meteor} \cite{banerjee-lavie-2005-meteor} metrics. We report two different results for each model, one on the original, and one for the cleaned set.

\begin{table}[ht]
	\centering
	\small
	\begin{tabular}{lll}
		\toprule[1.5pt]
		           &  \textsc{Rouge-L} & \textsc{Meteor} \\ \midrule
		\cite{scialom-etal-2020-mlsum} & 32.90/ \ \ -- & 26.30/ \ \ -- \\ \midrule
		\textsc{TrBart} (120M)    & 35.54/35.08   & 26.47/25.81   \\
		\textsc{mBart50} (680M)  & 39.21/38.47   & 30.84/30.36   \\
		\textsc{mT5-Base} (220M) & \textbf{39.92/38.76} & \textbf{31.72/31.47} \\ \bottomrule[1.5pt]
	\end{tabular}
	\caption{Evaluation of different models on \textsc{Mlsum} test set along with their no of parameters. Metrics are calculated for both (Original/Cleaned) test sets.}
	\label{tab:sum-results}
	\vspace{-15pt}
\end{table}

\vspace{\subsubsectionspacing}
\textbf{Baselines \ \ } 
As a baseline model for summarization, we present \textsc{TrBart}, a Seq2Seq Transformer \cite{vaswani-etal-2017-attention} trained following the configuration of BART Base \cite{lewis-etal-2020-bart}, which is a state-of-the-art model for abstractive summarization in English.

Moreover, we fine-tune two different pre-trained models. The first model is Multilingual BART (\textsc{mBart50}) \cite{tang-etal-2020-mbart50}, which is pre-trained on data from 50 different languages. The second model is Multilingual Text to Text Transformer (\textsc{mT5-Base}) \cite{xue-etal-2021-mt5}. As shown in Table \ref{tab:sum-results}, all models perform better than the best proposed baseline \cite{scialom-etal-2020-mlsum}, which follows the UniLM architecture \cite{dong-et-al-unilm}.

\subsection{Text Classification}
\label{subsec:textclassify}

Text classification can be utilized in several applications such as sentiment analysis or topic identification. In this task we take a sequence of text as an input, and output a probability distribution over the given classes. In our work on Turkish we benchmark three models on two datasets from different domains.

\vspace{\subsubsectionspacing}
\textbf{Datasets \ \ } 
We work on the news categorization (\textsc{News-Cat}) dataset \cite{amasyali-yildirim-2004-otomatik}. In this dataset, news articles are labeled with one of the following five categories \textit{health, sports, economy, politics, magazine}. There is no splits provided in the original work for \textsc{News-Cat} dataset. Hence we shuffle the dataset and construct our own splits in a stratified way, keeping the class distribution balanced across splits. We use 750 samples for training, 150 samples for validation, and 250 samples for testing. More details on the dataset can be found in Appendix \ref{app:cls}.

Since no information about the quality of annotation or Inter-annotator Agreement (IAA) rates were provided in for \textsc{News-Cat} \cite{amasyali-yildirim-2004-otomatik}, we applied a quality assessment by re-annotating the test set. We asked three annotators to label the documents of test set with one of the given five categories. The annotators agreed with the gold annotation with an average IAA rate of \textsc{Fleiss} $\kappa = 0.88$.

The second dataset is the corpus of Offensive Speech Identification in Social media (\textsc{Offenseval}) \cite{coltekin-2020-corpus}. This dataset was collected from Twitter, where the tweets are annotated for offensive speech with \textit{offensive}, or \textit{non-offensive} labels. We choose these datasets for benchmarking since they vary in domain and average article length.

\vspace{\subsubsectionspacing}
\textbf{Metrics \ \ } We use the macro averaged F1-Score to account for the imbalance in classes within the datasets.

\begin{table}[t]
\centering
\small
\begin{tabular}{lccc}
\toprule[1.5pt]
         & \textsc{Offenseval} & \textsc{News-Cat} & Avg. \\ \midrule
\textsc{BiLSTM}   & 0.747 & 0.808 & 0.777 \\
\textsc{CNN-Text} & 0.751 & 0.883 & 0.817 \\
\textsc{BerTURK}     & \textbf{0.823} & \textbf{0.944} & \textbf{0.883} \\ \bottomrule[1.5pt]
\end{tabular}
\caption{Evaluation results (macro averaged F1-Score) of our baseline models for text classification task. The last column represent the average F1-scores of each model.}
\label{tab:classify-results}
\vspace{-15pt}
\end{table}

\vspace{\subsubsectionspacing}
\textbf{Baselines \ \ } 
We measure the performance of three deep learning models—one with pre-training and two without pre-training. The pre-trained model is the \textsc{Bert} \cite{devlin-etal-2019-bert} based Turkish pre-trained (\textsc{BerTURK}) model \cite{schweter-2020-berturk}. The remaining two models employ randomly initialized word embeddings of size 256. In one of them we use two layers of Bidirectional LSTM (\textsc{BiLSTM}) \cite{sepp-schmidhuber-1997-lstm} with a hidden size of 256. In the other model (\textsc{CNN-Text}), we use Convolutional Neural Networks for Sentence Classification \cite{kim-2014-convolutional} with 32 filters instead of 2.

Looking at F1 scores in Table \ref{tab:classify-results}, we can observe the advantage of pre-trained \textsc{BerTURK} model over \textsc{BiLSTM} and \textsc{CNN-Text}.

%

  \section{Conclusion}
\label{sec:conclusion}

We believe that while some languages such as Turkish do not fall under the definition of under-resourced languages, they attract relatively little research interest as a result of the lack of organized benchmarks and baselines. To address this problem, we presented \textsc{Mukayese}, a comprehensive set of benchmarks along with corresponding baselines for seven different tasks: Language Modeling, Machine Translation, Named Entity Recognition, Sentence Segmentation, Spell Checking and Correction, Summarization, and Text Classification, as well as four new benchmarking datasets in Turkish for Language Modeling, Sentence Segmentation, and Spell Checking and Correction. For future work, the same methodology can be followed to include more tasks such as Dependency Parsing, Morphological Analysis, coreference resolution. We hope that \textsc{Mukayese} encourages more researchers to get involved in the development of Turkish NLP, and it sets an example and leads to an increase in efforts on under-researched languages.

\section*{Acknowledgements}

We thank Buse Çarık, Reyyan Yeniterzi, and Taner Sezer for their helpful discussions and feedback. Ali Safaya was supported by KUIS AI Center fellowship. Moreover, parts of the results reported in this paper were performed at TUBITAK ULAKBIM, High Performance and Grid Computing Center (TRUBA resources).

  \bibliography{bib/anthology,bib/custom}
  \bibliographystyle{other/acl_natbib}
  \appendix
\section{Turkish Language}
\label{app:turkish}


Even though in formal language the Subject-Object-Verb order is predominantly used, Turkish is a free-order language, meaning that words can freely change order depending on the context without changing the meaning but only the accentuation. The English sentence "I am going to school." can be translated into Turkish as "Ben okula gidiyorum." where all 6 permutations of the words are valid and meaningful:
\vspace{\subsubsectionspacing}

\begin{center}
\noindent\textit{
- Ben okula gidiyorum \\
- Ben gidiyorum okula \\
- Gidiyorum ben okula \\
- Gidiyorum okula ben \\
- Okula gidiyorum ben \\
- Okula ben gidiyorum \\
}\end{center}

In Turkish, morphologically ambiguous words are common in running texts. Depending on the context, the same word can have varying morphological features. For instance, the word "masalı" can correspond to the following:
\vspace{\subsubsectionspacing}

\begin{small}
\noindent
masal+Noun+A3sg+Pnon+Acc(=the story) \\
masal+Noun+A3sg+P3sg+Nom(=his story) \\
masa+Noun+A3sg+Pnon+NomˆDB+Adj+With(=with tables) \\ 
\end{small}

Given all these language features, Turkish language needs special attention by the research community, and we cannot assume that methods with good performance on English would yield good results on Turkish.

\section{Computational Costs and Implementations}

We utilize \textsc{Nvidia Tesla V100 GPUs} with 32GBs of memory for training our baselines. In Table \ref{tab:app-compute}, we depict approximate estimations of the training time for each of our compute-intensive baselines.

The implementations of \textsc{Transformer} \cite{vaswani-etal-2017-attention}, and \textsc{ConvS2S} \cite{gehring-etal-2017-convs2s} are based on the open-source library Fairseq \cite{ott-etal-2019-fairseq}. We use the Flair library \cite{akbik-etal-2019-flair} for the \textsc{Bert-crf} model in the Named Entity Recognition task. The remaining deep learning models used as our baselines are either implemented using the Huggingface Transformers library \cite{wolf-etal-2020-transformers}. All reported experiments and implementations of deep learning models are performed using PyTorch \cite{paszke-etal-2019-pytorch}. 

\begin{table}[!h]
\small
\begin{tabular}{llll}
\toprule[1.5pt]
\textbf{Model}               & \textbf{Dataset}        & \textbf{GPU Hr} & \textbf{Batch S.} \\
\midrule
\multicolumn{2}{l}{\textsc{Language Modeling}}  & & \\ & & & \\
SHA-RNN & trwiki-67 & 30 & 16 \\
SHA-RNN & trnews-64 & 24 & 32 \\
Adap. Transformer & trwiki-67 & 72 & 16 \\
Adap. Transformer & trnews-64 & 56 & 16 \\ \midrule
\multicolumn{2}{l}{\textsc{Machine Translation}}  & & \\ & & & \\
ConvS2S             & Wmt-16  & 12x2                      & 4000$^*$                  \\
ConvS2S             & MuST-C  & 11x2                      & 4000$^*$                 \\
Transformer         & Wmt-16  & 8x2                       & 4096$^*$                  \\
Transformer         & MuST-C  & 7x2                       & 4096$^*$                  \\
mBART50             & Wmt-16  & 24x2                      & 2                     \\
mBART50             & MuST-C  & 22x2                      & 2                     \\ \midrule
\multicolumn{2}{l}{\textsc{Summarization}} & & \\ & & & \\
Transformer         & Mlsum          & 12                        & 4                     \\
mBART50             & Mlsum          & 51                        & 2                     \\
mT5-Base            & Mlsum          & 38                        & 2                     \\
 \bottomrule[1.5pt]
\end{tabular}
\caption{Computational costs per models. $^*$ Fairseq uses dynamic batching, so we report max number of tokens per batch.}
\label{tab:app-compute}
\vspace{-5pt}
\end{table}

\section{Datasets and Baselines}
\subsection{Language Modeling}
\label{app:lm}

We provide some samples from \textsc{trwiki-67} and \textsc{trnews-64} corpora in Table \ref{tab:lm-samples}. These corpora are presented with minimal pre-processing. We remove non-Turkish characters and redundant texts such as category lists and tables from \textsc{trwiki-67}. Sentences and words are counted based on \texttt{sent\_tokenize}, \texttt{word\_tokenize} methods of NLTK \cite{bird-etal-2009-nltk}.

\begin{table}[!ht]
	\centering
	\small
	\begin{tabular}{lllll}
		\toprule[1.5pt]
		& \#articles & \#words & \#tokens & avg.sent \\ \midrule
		\multicolumn{2}{l}{\textsc{trwiki-67}}  &      &         &            \\
		Training           & 374K       & 63.5M & 139M & 12.8       \\
		Validation         & 10K        & 1.7M &  4M  & 13.3       \\
		Test               & 10K        & 1.7M &  4M & 12.9       \\ \midrule
		Total              & 394K       & 67M  &  147M   & 12.8       \\ \midrule
		\multicolumn{2}{l}{\textsc{trnews-64}}    &     & &            \\
		Training           & 140K       & 59.7M  & 421M & 23         \\
		Validation         & 5K         & 2.1M  & 15M  & 22.8       \\
		Test               & 5K         & 2.1M  &  15M & 22.9       \\ \midrule
		Total              & 150K       & 64M   & 450M  & 23         \\ \bottomrule[1.5pt]
	\end{tabular}
	\caption{Statistics about \textsc{trwiki-67} and \textsc{trnews-64} corpus splits. The column \textit{avg. sents} refers to the average number of sentences per article. Tokens are characters for \textsc{trnews-64} and sentencepiece tokens for \textsc{trwiki-67}.}
	\label{tab:lm-datasets}
	\vspace{-10pt}
\end{table}

We follow the same architectures proposed by \cite{merity-2019-sharnn, sukhbaatar-etal-2019-adaptive}. The only difference in architecture is based on vocabulary size due to difference in training data. For training, we use vocabulary size of 32K sentencepiece for \textsc{trwiki-67}, and 124 for \textsc{trnews-64} which includes the Turkish alphabet with punctuation and some other common characters. We train both models until no improvement over the validation set, then following the original implementation we lower the learning-rate, dividing it by 10 and run until no improvement on the validation set again.

\subsubsection{Normalizing perplexity}

The Perplexity metric is defined as the exponent of the average entropy over a corpus \cite{mikolov-etal-2011-perplexity}:
\begin{equation}
    \small
    \textsc{Ppl}(X_{test}) = exp(-\frac{1}{N} \sum_{i=1}^{n}log\ p_\theta(x_i|x_{test}{<i}))
\end{equation}
where $N$ is the original number of tokens in $X_{test}$, and $n$ is the number of tokens of $X_{test}$ when tokenized using a certain tokenization algorithm. Depending on what tokenization is used, $N$ might or might not be equal to $n$. To accommodate this issue, $N$ should always be the same when calculating perplexity for different models \cite{shoeybi-etal-2019-megatronlm}. 
\subsection{Named Entity Recognition (NER)}
\label{app:ner}

We provide statistics about dataset splits for both \textsc{Milliyet-Ner} and \textsc{WikiANN} in Table \ref{tab:ner-datasets}. 

\begin{table}[!ht]
\centering
\small
\begin{tabular}{llll}
\toprule[1.5pt]
             & Training  & Validation & Test  \\ \midrule
\textsc{WikiANN}       &      &      &       \\
Location     & 9679   & 5014 & 4914  \\ 
Organization & 7970   & 4129 & 4154  \\
Person       & 8833   & 4374 & 4519  \\ \midrule
Total words       & 149786 & 75930 & 75731 \\ \midrule[1.5pt]

\textsc{Milliyet-Ner}     &     &     & \\
Location     & 8821   &  942  & 1126  \\
Organization & 8316   &  842  & 873   \\
Person       & 13290  &  1400  & 1603  \\ \midrule
Total words  & 419996 &  45532  & 49595 \\ \toprule[1.5pt]

\end{tabular}
\caption{Distribution of Named entities over classes in \textsc{Milliyet-Ner} and \textsc{WikiANN} datasets.}
\label{tab:ner-datasets}
\vspace{-5pt}
\end{table}

\subsection{Machine Translation}
\label{app:mt}

We utilize two datasets for Machine Translation, \textsc{WMT-16} dataset, which was presented at the first Conference of Machine Translation (WMT), and MuST-C dataset.
This corpus was extracted from movies and TV shows subtitles.
Statistics of both datasets are presented in Table \ref{tab:mt-datasets}.

\begin{table}[h]
	\centering
    \small
	\begin{tabular}{lll}
		\toprule[1.5pt]
		            & \#Sentences & \#Words \\ \midrule
		\textit{Turkish} & & \\ 
		\textsc{Must-C}  & 236K / 1.3K / 2K   & 3.4M / 19K / 33K      \\
		\textsc{WMT-16}     & 205K / 1K / 3K    & 3.6M / 14K / 44K       \\ \midrule
		\textit{English} & & \\
		\textsc{Must-C}  & 236K / 1K/ 2K    & 4.6M / 26K / 45K      \\
		\textsc{WMT-16}     & 205K / 1K / 3K    & 4.4M / 19K / 58K       \\
         \bottomrule[1.5pt]
	\end{tabular}
	\caption{Statistics of machine translation datasets. Each cell represents the (Train / Validation / Test) values of the datasets in the corresponding row. \textsc{Wmt-16} and \textsc{MuST-C} refer to Turkish-English subsets.}
	\label{tab:mt-datasets}
	\vspace{-5pt}
\end{table}

\subsection{Sentence Segmentation}
\label{app:sentenceseg}

In this section, we provide additional information for our Sentence Segmentation \ref{subsec:sent-segmentation} Benchmark. 

In both clean and corrupted training cases, ErSatz and Punkt are trained with all subsets. Following the authors, our baseline model ErSatz is trained without changing the original architecture with a vocabulary size of 500, left and right context of 5 for 100 epochs using early stopping. We use the NLTK \cite{bird-etal-2009-nltk} implementation of the Punkt tokenizer \cite{kiss-strunk-2006-unsupervised} for both training and testing purposes. The spaCy tokenizer \cite{montani-etal-2021-spacy} is used with the default settings provided by the library. 

\begin{table}[h!]
	\centering
	\small
	\begin{tabular}{llll}
        \toprule[1.5pt]
                  & \#Articles & \#Sentences & \#Words \\ \midrule
        News      & 300 & 6K & 102K \\
        Tweets    & 10K & 28K & 242K \\
        Abstracts & 300 & 6K & 112K \\ \midrule
        Total     & 10.6K & 40K & 456K \\ \bottomrule[1.5pt]
    \end{tabular}
	\caption{Statistics of \textsc{trseg-41} dataset.}
	\label{tab:segm-datasets}
	\vspace{-15pt}
\end{table}

Table \ref{tab:sentenceseg-samples} provides examples from each subset of the \textsc{trseg-41} dataset along with their corrupted versions. The dataset is annotated by a single human. The reason for maximizing the number of abbreviations and proper nouns is that rule-based methods are designed to be sensitive to local language features such as periods and capital letters. In editorial texts, sentence segmentation can achieve high success. Therefore, we apply automated random corruption process as described in Section \ref{subsec:sent-segmentation}. The rationale behind this is to eliminate the aforementioned context for rule-based approaches and to promote learning methods. 

Table \ref{tab:sentenceseg-preds} shows examples of the results of our baselines. The results show that while the models are able to perform successful sentence segmentation on clean editorial text, they experience an evident drop in performance on corrupted versions.

\subsubsection{F1-Score}
In this benchmark, we compare the performances of the models via F1-Score. For sentence segmentation, we define F1-Score as the accuracy measure of the position of the dots in a given piece of text among spaced tokens. This means that for a paragraph containing $N$ words, all words are separated as distinct tokens, leaving $N - 1$ locations to place the dots as separators. Our measure is based on the correctness of the placed dots in this given setting. We calculate F1-Score in the following way:
\begin{equation}
    \small
\frac{TP}{TP + 1/2(FP+FN)}
\end{equation}

where TP is true positive rate, FP is false positive rate, and FN is false negative rate. Our calculation is based on the \textit{Scorer} submodule of the spaCy library.

\subsection{Spellchecking and Correction}
\label{app:spellcheck}

In this section, we provide a detailed description of the spellchecking dataset with the statistics about the word set and corruption methods.

The dataset consist of 10K words it total, and includes pairs of gold and corrupted words. 8500 words are randomly sampled from TS Corpus Word List \cite{sezer-2013-tscorpus,sezer-2017-tscorpus}, 1K random words are included from foreign language and 500 randomly generated word-like character sequences are added.

For 70\% of the sampled Turkish words, we apply one corruption with 70\% probability, two corruptions with 25\% probability and three corruptions with 5\% probability. The following corruption methods with their probability distribution is applied for a single corruption:

\begin{itemize}
  \item For a probability of 1/2, the word is asciified.
  \item For a probability of 1/6, a random character in the word is substituted by another character sampled from a distribution simulating the placement of keys in standard Turkish-Qwerty keyboards.
  \item For a probability of 1/6, a random character is inserted into the word sampled from a distribution simulating the placement of keys in standard Turkish-Qwerty keyboards.
  \item For a probability of 1/6, a random character is deleted from a word sampled from a distribution simulating the placement of keys in standard Turkish-Qwerty keyboards.
\end{itemize}

The remaining 30\% of the words are uncorrupted, therefore their gold and input versions are same. For evaluating against inserted foreign words and randomly generated character sequences where no gold output exists, we use an empty string as the gold output.
\subsection{Summarization}
\label{app:sum}

We remove these instances from the dataset for a more accurate evaluation and evaluate our models on both the original and the cleaned sets. In Table \ref{tab:sum-datasets}, we provide some statistics about both sets, before and after the deduplication.

\begin{table}[h!]
	\centering
	\small
	\begin{tabular}{lcc}
		\toprule[1.5pt]
		& Original & Cleaned \\ \midrule
		Avg. article length   & 259.1  & 258.4   \\
		Avg. summary length   & 18.5   & 18.3    \\ \midrule
		\textit{Splits}     &          &     \\
		Training   & 249277   & 246490       \\
		Validation & 11565    & 10852        \\
		Test       & 12775    & 11897        \\ \midrule		Total      & 273617   & 269239       \\ \bottomrule[1.5pt]
	\end{tabular}
	\caption{Statistics of the Turkish subset of \textsc{Mlsum}. The number of samples is provided for each split before and after the deduplication.}
	\label{tab:sum-datasets}
	\vspace{-10pt}
\end{table}

We provide summaries predicted by our models in Table \ref{tab:sum-samples}.
\subsection{Text Classification}
\label{app:cls}

In table \ref{tab:classification-datasets}, we provided statistics about both of the datasets we used for text classification task. 

\begin{table}[h!]
	\centering
	\small
	\begin{tabular}{lll}
		\toprule[1.5pt]
		& \textsc{Offenseval} & \textsc{News-Cat} \\ \midrule
		Avg. \#words   & 8.5  & 227.3   \\
	    \#Classes   & 2   & 5    \\ \midrule
		\textit{Splits}     &          &     \\
		Training   & 28000   & 750       \\
		Validation & 3277    & 150        \\
		Test       & 3515    & 250        \\ \midrule
		Total      & 34792   & 1150       \\ \bottomrule[1.5pt]
	\end{tabular}
	\caption{Statistics of \textsc{News-Cat} and \textsc{Offenseval} dataset splits.}
	\label{tab:classification-datasets}
	\vspace{-15pt}
\end{table}

\begin{table*}
\scriptsize
\begin{tabulary}{\textwidth}{L}
\toprule
\textsc{trwiki-67} \\
\midrule
== NGC 1710 == \\
\\
NGC 1710, Yeni Genel Katalog'da yer alan bir galaksidir. Gökyüzünde Aslan takımyıldızı yönünde bulunur. E-S0 tipi bir merceksi, eliptik galaksidir. Amerikan astronom Francis Leavenworth tarafından 1885 yılında 66,04 cm (26 inç) çaplı mercekli tip bir teleskopla keşfedilmiştir. \\
\\
== Şenol Gürşan == \\
\\
Şenol Gürşan, (d. 17 Ekim 1964, Pınarhisar, Kırklareli) Türk avukat ve siyasetçi.\\
İstanbul Üniversitesi Hukuk Fakültesi'ni bitirmiş ve serbest avukat olarak çalışmıştır. Kırklareli İlim Yayma Cemiyeti Kuruculuğu ve Başkanlığı görevlerinde bulunmuştur.\\
2009 yılında Adalet ve Kalkınma Partisi Kırklareli il yönetim kurulu üyesi olmuş, TBMM 24. dönem AK Parti Kırklareli milletvekili, Türkiye-Polonya Dostluk Grubu Başkanı ve TBMM KİT Komisyonu Sözcüsü olmuştur. Gelecek Partisi Kurucular Kurulu üyesi olup aynı zamanda partinin genel sekreteridir.\\
İyi düzeyde Almanca bilen Gürşan, evli ve 2 çocuk babasıdır. \\
\midrule
\textsc{trnews-64} \\
\midrule
Dolar dün 2.5075 liraya kadar çıkarak rekor kırmasının ordından bugün 2.49 - 2.50 lira aralığında hareket etti. Cari işlemler açığının beklentilere paralel gelmesinin de etkisiyle 2.4820 liraya kadar çekilen dolar, daha sonra gelens alımlarla 2.5085'e çıkarak rekorunu tazeledi. ABD para birimi daha sonra 2.5050 - 2.5070 düzeylerinde hareket ederken, euro da 2.8380 lira düzeylerine çıktı ve yarı yarıya euro ve dolardan oluşan döviz sepeti de 2.63 düzeyinin üstüne çıktı. \\
\\
DW Türkçe Servisi’nin aktardığına göre, ‘Aghet’ (Ağıt) konserinin Almanya’nın İstanbul Başkonsolosluğu’ndaki temsiline Cumhurbaşkanı Recep Tayyip Erdoğan da davet edildi. Alman haber ajansı dpa’nın haberinde, Erdoğan’ın yanı sıra Başbakan Binali Yıldırım, Dışişleri Bakanı Mevlüt Çavuşoğlu ile Kültür ve Turizm Bakanı Nabi Avcı’nın da davetliler arasında olduğu belirtildi. Habere göre, gönderilen davetiyelerde etkinlikte ‘Türk ve Ermeni geçmişlerindeki yaralar’ ile ifade ve sanat özgürlüğünün ele alınacağı ifade edildi. Dresden Senfoni Orkestrası tarafından hazırlanan ‘Aghet’ konseri, İstanbul Başkonsolosluğu’nda 13 Kasım’da gerçekleştirilecek. Etkinlikte ayrıca Türk-Ermeni-Alman Dostluk Derneği’nin kurulması planlanıyor. \\
\bottomrule
\end{tabulary}
\caption{Text samples from \textsc{trwiki-67} and \textsc{trnews-64} corpora.}
\label{tab:lm-samples}
\end{table*}

\begin{table*}
\scriptsize
\begin{tabulary}{\textwidth}{L}
\toprule
\textsc{Input} \\
\midrule
Bursa İnegöl ilçesi Deydinler Mahallesi'nde yaşayan Erdoğan Bitirim evde gördüğü yılanı elleriyle yakalayıp doğaya saldı. Havaların sıcak olmasıyla birlikte son günlerde sayıları artan yılanlar vatandaşları tedirgin ediyor. Erdoğan Bitirim evinde yakaladığı yılanı doğaya salarken o anları cep telefonuyla kayıt altına aldı. Bitirim, yılana her hangi bir zarar vermediği belirterek, "Çok hızlı ve seri hareket ediyordu. Birkaç kez bana saldırmaya kalktı ama ben onu yakaladım. Yakaladığım yılanı zarar vermeden doğa saldım. Yaklaşık 1 metre boyunda bir yılandı" dedi. \\
\midrule
\textsc{Reference} \\
\midrule
Bursa'nın İnegöl ilçesinde bir vatandaş evinde eliyle yakaladığı yılanı doğaya saldı. \\
\midrule
\textsc{TrBart} \\
\midrule
bursa'nın inegöl ilçesinde yaşayan erdoğan bitirim, yılanı elleriyle yakalayıp doğaya saldı.\\
\midrule
\textsc{mBart50} \\
\midrule
BURSA'nın İnegöl ilçesinde yaşayan Erdoğan Bitirim, evde gördüğü yılanı elleriyle yakalayıp doğaya saldı.\\
\midrule
\textsc{mT5-Base} \\
\midrule
Bursa'nın İnegöl ilçesinde yaşayan Erdoğan Bitirim evinde yakaladığı yılanı doğaya saldı.\\
\bottomrule
\end{tabulary}
\caption{Example of summaries generated by the three baselines for a sample from the test set of \textsc{Mlsum}.}
\label{tab:sum-samples}
\end{table*}

\begin{table*}[t]
\scriptsize
\begin{tabulary}{\textwidth}{L}
\toprule
\textbf{Clean Abstract Sample:} \\
\midrule
Bu çalışmanın amacı, bayan ve erkek voleybolcular ile güreşçilerin statik, yaylanarak, düşerek ve tekrarlı sıçrama performanslarını karşılaştırmaktır.\\
Bu çalışmaya Yaşar Doğu Beden Eğitimi ve Spor Yüksekokulunda okuyan 2. ve 3. Ligde mücadele eden 20 bayan voleybolcu, 20 erkek voleybolcu ile Milli 20 erkek güreşçi gönüllü olarak katılmıştır.\\
Bayan voleybolcuların yaş ortalaması 21.15 yıl, voleybolcu erkeklerin 20.80 yıl ve güreşçilerin 20.60 yıldır.\\
Bütün denekler statik sıçrama, yaylanarak sıçrama, düşerek sıçrama ve tekrarlı sıçrama yapmışlardır.\\
Sıçrama değerlerinin belirlenmesi, New Test Power Timer System 300 Series aleti kullanılarak yapılmıştır.\\
Ayrıca çalışmaya katılan sporcuların, beden kitle indeksi (BKİ), esneklik ve vücut yağ yüzdesi değerleri ölçülmüştür.\\
Üç grup arasında fark olup olmadığına bakmak amacıyla Kruskal Vallis testi, ikili karşılaştırmalarda Mann Whitney U testi kullanılmıştır.\\
Sporcuların karşılaştırıldığında güreşçi erkeklerin voleybolcu erkelerden daha esnek oldukları görülmüştür.\\
boy, vücut ağırlığı, BKI, vücut yağ yüzdesi arasında anlamlı derecede farklılık bulunmuştur.\\
Voleybolcu erkeklerin Düşerek, Statik, Yaylanarak ve Tekrarlı sıçrama yükseklikleri ve güçleri voleybolcu bayanlardan ve güreşçilerden yüksek bulunmuştur.\\
Güreşçilerin ise statik ve yaylanarak sıçrama yükseklikleri ve güçleri bayan voleybolculardan daha yüksek bulunmuştur.\\
Erkek voleybolcuların sıçrama değerlerinin güreşçilerden yüksek çıkması yapılan spor branşı ile ilgilidir.\\
Voleybolcu bayanların sıçrama performansının güreşçi erkeklerden daha iyi olması beklenirken cinsiyet faktörünün bu durumun önüne geçtiği görülmüştür.\\
Sonuç olarak, yapılan spor branşının ve cinsiyetin sıçrama performansı üzerinde önemli etkisinin olduğu görülmüştür.\\
\midrule[1pt]
\textbf{Corrupted Abstract Sample:} \\
\midrule
BU ÇALIŞMANIN AMACI BAYAN VE ERKEK VOLEYBOLCULAR İLE GÜREŞÇİLERİN STATİK YAYLANARAK DÜŞEREK VE TEKRARLI SIÇRAMA PERFORMANSLARINI KARŞILAŞTIRMAKTIR\\
bu çalışmaya yaşar doğu beden eğitimi ve spor yüksekokulunda okuyan 2. ve 3. ligde mücadele eden 20 bayan voleybolcu, 20 erkek voleybolcu ile milli 20 erkek güreşçi gönüllü olarak katılmıştır.\\
Bayan voleybolcuların yaş ortalaması 21.15 yıl, voleybolcu erkeklerin 20.80 yıl ve güreşçilerin 20.60 yıldır.\\
Bütün denekler statik sıçrama yaylanarak sıçrama düşerek sıçrama ve tekrarlı sıçrama yapmışlardır\\
Sıçrama değerlerinin belirlenmesi, New Test Power Timer System 300 Series aleti kullanılarak yapılmıştır.\\
Ayrıca çalışmaya katılan sporcuların beden kitle indeksi BKİ esneklik ve vücut yağ yüzdesi değerleri ölçülmüştür\\
üç grup arasında fark olup olmadığına bakmak amacıyla kruskal vallis testi, ikili karşılaştırmalarda mann whitney u testi kullanılmıştır.\\
Sporcuların karşılaştırıldığında güreşçi erkeklerin voleybolcu erkelerden daha esnek oldukları görülmüştür.\\
boy, vücut ağırlığı, BKI, vücut yağ yüzdesi arasında anlamlı derecede farklılık bulunmuştur.\\
Voleybolcu erkeklerin Düşerek Statik Yaylanarak ve Tekrarlı sıçrama yükseklikleri ve güçleri voleybolcu bayanlardan ve güreşçilerden yüksek bulunmuştur\\
Güreşçilerin ise statik ve yaylanarak sıçrama yükseklikleri ve güçleri bayan voleybolculardan daha yüksek bulunmuştur\\
Erkek voleybolcuların sıçrama değerlerinin güreşçilerden yüksek çıkması yapılan spor branşı ile ilgilidir\\
VOLEYBOLCU BAYANLARIN SIÇRAMA PERFORMANSININ GÜREŞÇİ ERKEKLERDEN DAHA İYİ OLMASI BEKLENİRKEN CİNSİYET FAKTÖRÜNÜN BU DURUMUN ÖNÜNE GEÇTİĞİ GÖRÜLMÜŞTÜR\\
Sonuç olarak, yapılan spor branşının ve cinsiyetin sıçrama performansı üzerinde önemli etkisinin olduğu görülmüştür.\\
\midrule[1pt]
\textbf{Clean Tweet Sample:} \\
\midrule
@user @user o kulların açılmasını 1 gün erteledi, çünkü alın size müjde veriyorum diyecek.\\
başka bişey yok, işler çığırından çıkmış\\
\midrule[1pt]
\textbf{Corrupted Tweet Sample:} \\
\midrule
@user @user     O KULLARIN AÇILMASINI 1 GÜN ERTELEDİ ÇÜNKÜ ALIN SİZE MÜJDE VERİYORUM DİYECEK\\
BAŞKA BİŞEY YOK İŞLER ÇIĞIRINDAN ÇIKMIŞ\\
\midrule[1pt]
\textbf{Clean News Sample:} \\
\midrule
Kanal 2 televizyonunda, İsrail'in tanınmış gazeteci ve analistlerinden Ehud Yaari ile birlikte konuk olan Lieberman'a, "Bakanlığı döneminde hem Türkiye hem de Mısır'dan büyükelçilerin kovulduğuna" işaret edilerek, gazetede yayımlanan haberin doğru olup olmadığı soruldu.\\
Haberin doğru olmadığını söyleyen lieberman'a, bu kez, "Peki (Dışişleri'nde) böyle şeyler konuştunuz mu?" sorusu yöneltildi.\\
Lieberman, bu soruya, "Dışişleri Bakanlığı'nda her gün yüzlerce fikir tartışma konusu edilir" yanıtını verdi.\\
Bunun üzerine, Ehud Yaari, "Açıkça söyleyin, PKK terör örgütüne silah vs. sağlama gibi, yardım etme konusu konuşuldu mu?" diyerek, sorusunu yineledi.\\
Lieberman, soruya bu kez "Hayır, kesinlikle konuşulmadı" karşılığını verdi.\\
Lieberman, Palmer Komisyonu raporunun "Mavi Marmara" baskını ile ilgili olarak İsrail'in eyleminin ve Gazze'ye ablukanın haklı olduğunu açıkça ortaya koyduğunu da ifade etti.\\
Lieberman, Türkiye ile ilişkilerin normalleştirilmesinin yeniden sağlanacağı ve Türkiye'nin, böyle bir normalleşmenin çıkarına olacağını göreceği umudunda olduğunu da kaydetti.\\
"Alevlerin seviyesini düşürmeye çalışıyoruz" İsrail Başbakanı Binyamin Netanyahu da Türkiye ile yaşanan krizin kendi seçimleri olmadığını öne sürdü.\\
Türkiye ile ilişkilerin daha da kötüye gitmesini önlemeye çalıştıklarını savunan Netanyahu, halihazırda, iki ülke arasındaki "Alevlerin seviyesini düşürmeye" uğraştıklarını belirterek, "Umarım bu gerginlik, eğer karşı taraf da isterse, sona erdirilecektir" diye konuştu.\\
\midrule[1pt]
\textbf{Corrupted News Sample:} \\
\midrule
Kanal 2 televizyonunda, İsrail'in tanınmış gazeteci ve analistlerinden Ehud Yaari ile birlikte konuk olan Lieberman'a, "Bakanlığı döneminde hem Türkiye hem de Mısır'dan büyükelçilerin kovulduğuna" işaret edilerek, gazetede yayımlanan haberin doğru olup olmadığı soruldu.\\
Haberin doğru olmadığını söyleyen liebermana bu kez Peki Dışişlerinde böyle şeyler konuştunuz mu sorusu yöneltildi\\
Lieberman, bu soruya, "Dışişleri Bakanlığı'nda her gün yüzlerce fikir tartışma konusu edilir" yanıtını verdi.\\
Bunun üzerine Ehud Yaari Açıkça söyleyin PKK terör örgütüne silah vs sağlama gibi yardım etme konusu konuşuldu mu diyerek sorusunu yineledi\\
LIEBERMAN, SORUYA BU KEZ "HAYIR, KESİNLİKLE KONUŞULMADI" KARŞILIĞINI VERDİ.\\
lieberman palmer komisyonu raporunun mavi marmara baskını ile ilgili olarak israilin eyleminin ve gazzeye ablukanın haklı olduğunu açıkça ortaya koyduğunu da ifade etti\\
lieberman, türkiye ile ilişkilerin normalleştirilmesinin yeniden sağlanacağı ve türkiye'nin, böyle bir normalleşmenin çıkarına olacağını göreceği umudunda olduğunu da kaydetti.\\
Alevlerin seviyesini düşürmeye çalışıyoruz İsrail Başbakanı Binyamin Netanyahu da Türkiye ile yaşanan krizin kendi seçimleri olmadığını öne sürdü\\
Türkiye ile ilişkilerin daha da kötüye gitmesini önlemeye çalıştıklarını savunan Netanyahu halihazırda iki ülke arasındaki Alevlerin seviyesini düşürmeye uğraştıklarını belirterek Umarım bu gerginlik eğer karşı taraf da isterse sona erdirilecektir diye konuştu \\
\bottomrule
\end{tabulary}
\caption{A sample from each of the abstracts, news, and tweets test subsets of \textsc{trseg-41}. \textbf{Clean} means the unedited and uncorrupted version of the data. \textbf{Corrupted} is the corrupted version of this abstract as specified in Section \ref{subsec:sent-segmentation}. The annotation of each sample is denoted by line-separation.}
\label{tab:sentenceseg-samples}
\end{table*}


\begin{table*}[t]
\scriptsize
\begin{tabulary}{\textwidth}{L}
\toprule
\textbf{Punkt Tokenizer Corrupted Tweet Sample Output:}\\
\midrule
@user @user     O KULLARIN AÇILMASINI 1 GÜN ERTELEDI ÇÜNKÜ ALIN SIZE MÜJDE VERIYORUM DIYECEK BAŞKA BIŞEY YOK IŞLER ÇIĞIRINDAN ÇIKMIŞ\\
\end{tabulary}
\begin{tabulary}{\textwidth}{L}
\midrule[1pt]
\textbf{spaCy Tokenizer Corrupted Tweet Sample Output:}\\
\midrule
@user @user     O KULLARIN AÇILMASINI 1 GÜN ERTELEDI ÇÜNKÜ ALIN SIZE MÜJDE VERIYORUM DIYECEK BAŞKA BIŞEY YOK IŞLER ÇIĞIRINDAN ÇIKMIŞ\\
\end{tabulary}
\begin{tabulary}{\textwidth}{L}
\midrule[1pt]
\textbf{ErSatz Tokenizer Corrupted Tweet Sample Output:}\\
\midrule
@user @user     O KULLARIN AÇILMASINI 1 GÜN ERTELEDI ÇÜNKÜ ALIN SIZE MÜJDE VERIYORUM DIYECEK BAŞKA BIŞEY YOK IŞLER ÇIĞIRINDAN ÇIKMIŞ\\
\end{tabulary}
\begin{tabulary}{\textwidth}{L}
\midrule[1pt]
\textbf{Punkt Tokenizer Corrupted News Sample Output:}\\
\midrule
Kanal 2 televizyonunda, İsrail'in tanınmış gazeteci ve analistlerinden Ehud Yaari ile birlikte konuk olan Lieberman'a, "Bakanlığı döneminde hem Türkiye hem de Mısır'dan büyükelçilerin kovulduğuna" işaret edilerek, gazetede yayımlanan haberin doğru olup olmadığı soruldu.\\
Haberin doğru olmadığını söyleyen liebermana bu kez Peki Dışişlerinde böyle şeyler konuştunuz mu sorusu yöneltildi\\
Lieberman, bu soruya, "Dışişleri Bakanlığı'nda her gün yüzlerce fikir tartışma konusu edilir" yanıtını verdi.\\
Bunun üzerine Ehud Yaari Açıkça söyleyin PKK terör örgütüne silah vs sağlama gibi yardım etme konusu konuşuldu mu diyerek sorusunu yineledi LIEBERMAN, SORUYA BU KEZ "HAYIR, KESİNLİKLE KONUŞULMADI" KARŞILIĞINI VERDİ.\\
lieberman palmer komisyonu raporunun mavi marmara baskını ile ilgili olarak imidrulesrailin eyleminin ve gazzeye ablukanın haklı olduğunu açıkça ortaya koyduğunu da ifade etti lieberman, türkiye ile ilişkilerin normalleştirilmesinin yeniden sağlanacağı ve türkiye'nin, böyle bir normalleşmenin çıkarına olacağını göreceği umudunda olduğunu da kaydetti.\\
Alevlerin seviyesini düşürmeye çalışıyoruz İsrail Başbakanı Binyamin Netanyahu da Türkiye ile yaşanan krizin kendi seçimleri olmadığını öne sürdü Türkiye ile ilişkilerin daha da kötüye gitmesini önlemeye çalıştıklarını savunan Netanyahu halihazırda iki ülke arasındaki Alevlerin seviyesini düşürmeye uğraştıklarını belirterek Umarım bu gerginlik eğer karşı taraf da isterse sona erdirilecektir diye konuştu\\
\end{tabulary}
\begin{tabulary}{\textwidth}{L}
\midrule[1pt]
\textbf{spaCy Tokenizer Corrupted News Sample Output:}\\
\midrule
Kanal 2 televizyonunda, İsrail'in tanınmış gazeteci ve analistlerinden Ehud Yaari ile birlikte konuk olan Lieberman'a, "Bakanlığı döneminde hem Türkiye hem de Mısır'dan büyükelçilerin kovulduğuna" işaret edilerek, gazetede yayımlanan haberin doğru olup olmadığı soruldu.\\
Haberin doğru olmadığını söyleyen liebermana bu kez Peki Dışişlerinde böyle şeyler konuştunuz mu sorusu yöneltildi Lieberman, bu soruya, "Dışişleri Bakanlığı'nda her gün yüzlerce fikir tartışma konusu edilir" yanıtını verdi.\\
Bunun üzerine Ehud Yaari Açıkça söyleyin PKK terör örgütüne silah vs sağlama gibi yardım etme konusu konuşuldu mu diyerek sorusunu yineledi LIEBERMAN, SORUYA BU KEZ "HAYIR, KESİNLİKLE KONUŞULMADI" KARŞILIĞINI VERDİ.\\
lieberman palmer komisyonu raporunun mavi marmara baskını ile ilgili olarak israilin eyleminin ve gazzeye ablukanın haklı olduğunu açıkça ortaya koyduğunu da ifade etti lieberman, türkiye ile ilişkilerin normalleştirilmesinin yeniden sağlanacağı ve türkiye'nin, böyle bir normalleşmenin çıkarına olacağını göreceği umudunda olduğunu da kaydetti.\\
Alevlerin seviyesini düşürmeye çalışıyoruz İsrail Başbakanı Binyamin Netanyahu da Türkiye ile yaşanan krizin kendi seçimleri olmadığını öne sürdü Türkiye ile ilişkilerin daha da kötüye gitmesini önlemeye çalıştıklarını savunan Netanyahu halihazırda iki ülke arasındaki Alevlerin seviyesini düşürmeye uğraştıklarını belirterek Umarım bu gerginlik eğer karşı taraf da isterse sona erdirilecektir diye konuştu \\
\bottomrule
\end{tabulary}

\begin{tabulary}{\textwidth}{L}
\textbf{ErSatz Tokenizer Corrupted News Sample Output:} \\
\midrule
Kanal 2 televizyonunda, İsrail'in tanınmış gazeteci ve analistlerinden Ehud Yaari ile birlikte konuk olan Lieberman'a, "Bakanlığı döneminde hem Türkiye hem de Mısır'dan büyükelçilerin kovulduğuna" işaret edilerek, gazetede yayımlanan haberin doğru olup olmadığı soruldu.\\
Haberin doğru olmadığını söyleyen liebermana bu kez Peki Dışişlerinde böyle şeyler konuştunuz mu sorusu yöneltildi\\
Lieberman, bu soruya, "Dışişleri Bakanlığı'nda her gün yüzlerce fikir tartışma konusu edilir" yanıtını verdi.\\
Bunun üzerine Ehud Yaari Açıkça söyleyin PKK terör örgütüne silah vs sağlama gibi yardım etme konusu konuşuldu mu diyerek sorusunu yineledi\\
LIEBERMAN, SORUYA BU KEZ "HAYIR, KESİNLİKLE KONUŞULMADI" KARŞILIĞINI VERDİ.\\
lieberman palmer komisyonu raporunun mavi marmara baskını ile ilgili olarak israilin eyleminin ve gazzeye ablukanın haklı olduğunu açıkça ortaya koyduğunu da ifade etti lieberman, türkiye ile ilişkilerin normalleştirilmesinin yeniden sağlanacağı ve türkiye'nin, böyle bir normalleşmenin çıkarına olacağını göreceği umudunda olduğunu da kaydetti.\\
Alevlerin seviyesini düşürmeye çalışıyoruz İsrail Başbakanı Binyamin Netanyahu da Türkiye ile yaşanan krizin kendi seçimleri olmadığını öne sürdü Türkiye ile ilişkilerin daha da kötüye gitmesini önlemeye çalıştıklarını savunan Netanyahu halihazırda iki ülke arasındaki Alevlerin seviyesini düşürmeye uğraştıklarını belirterek Umarım bu gerginlik eğer karşı taraf da isterse sona erdirilecektir diye konuştu\\
\bottomrule
\end{tabulary}
\caption{Predictions of the proposed ErSatz, Punkt, and spaCy baselines. ErSatz and Punkt are trained on the \textbf{Clean} version of the \textsc{trseg-41} training set. The listed predictions are for the samples provided in Table \ref{tab:sentenceseg-samples}.}
\label{tab:sentenceseg-preds}
\end{table*}



\end{document}